\documentclass[conference]{IEEEtran}

\usepackage{cite}
\usepackage{amsmath,amssymb,amsfonts}
\usepackage{graphicx}
\usepackage{textcomp}
\usepackage{xcolor}
\usepackage{titlesec}

\usepackage{tikz}
\usepackage{pgfplots}
\pgfplotsset{compat=1.18}

\usepackage[ruled,vlined,linesnumbered]{algorithm2e}
\usepackage[hidelinks]{hyperref}

\SetKwInput{KwObjective}{Objective}
\SetKwInput{KwInput}{Input}
\SetKwInput{KwOutput}{Output}
\SetKw{KwRet}{return}
\usepackage{booktabs}
\begin{document}

\title{SpatialVec: A Unified Representation Learning Framework for
Geospatial Objects with Boundary-Aware Sampling \\\Large Accepted to 2026 IEEE International Conference on Big Data}


\author{
\IEEEauthorblockN{Samya Acharja}
\IEEEauthorblockA{
\textit{Marquette University} \\
Milwaukee, USA \\
samya.acharja@marquette.edu
}
\and
\IEEEauthorblockN{Kanchan Chowdhury}
\IEEEauthorblockA{
\textit{Marquette University} \\
Milwaukee, USA \\
kanchan.chowdhury@marquette.edu
}
}

\maketitle

\begin{abstract}
    Spatial representation learning is essential for Geospatial Artificial Intelligence (GeoAI) applications, enabling machine learning models to capture the geometry, location, and spatial relationships of geographic entities such as points, polylines, and polygons. 
    Existing spatial encoding methods often struggle to preserve fine-grained topological information, particularly for interactions involving complex geometries such as polylines and polygons. These approaches inadequately capture boundary structure and local spatial context, causing topology-sensitive relationships near geometry boundaries to be underrepresented in the learned embedding space. Consequently, accurately modeling complex spatial interactions remains a significant challenge for downstream spatial reasoning tasks.

    
    To fill the gap, we introduce SpatialVec, a boundary-aware spatial representation learning framework that adaptively samples geometry boundaries and their surrounding regions, capturing fine-grained geometric structures critical for topology-aware spatial reasoning. A transformer-based encoder-decoder trained through geometry reconstruction produces compact, geometry-aware, and unified representations for points, polylines, and polygons. 
    Empirical evaluation on two real-world datasets shows that SpatialVec substantially improves the preservation of challenging topological relationships involving polygonal interactions and outperforms state-of-the-art baselines \textcolor{black}{significantly on downstream prediction and classification tasks}.

\end{abstract}

\section{Introduction}
Spatial representation learning has become an important foundation for
Geospatial Artificial Intelligence (GeoAI). Many GeoAI applications require
machine learning models to reason over geographic entities represented as
points, polylines, and polygons. Unlike images or text, these objects carry
geometric structure, absolute location, scale, orientation, and relationships
with neighboring entities. A useful spatial representation must therefore
retain this information to support tasks such as urban analysis,
transportation modeling, environmental monitoring, and spatial search
\cite{appli_urban,appli_trans,appli_weather,appli_spaitalreason,
appli_spaitalreason2,burns2026community}.

Several methods have been developed to encode spatial objects into fixed-dimensional vectors. Histogram-based spatial embeddings \cite{hist2vec} summarize the distribution of spatial data for downstream learning and query optimization. Poly2Vec \cite{poly2vec} provides a unified encoder for points, polylines, and polygons by representing geometric components in the Fourier domain \cite{poly2vec}. Geo2Vec\cite{geo2vec} instead samples a signed distance field around each object and learns a representation that captures shape and location. These methods show that spatial embeddings can replace raw coordinates in several downstream tasks providing a common interface for machine learning models.

However, accurately preserving spatial relationships between geometries remains difficult. Topological relationships such as touching, intersection, overlap, and containment are often determined by small regions where two boundaries meet or pass close to one another. Representations based mainly on global statistics \cite{hist2vec} or frequency-domain summaries \cite{poly2vec} may retain the overall geometry while weakening these local interactions. \textcolor{black}{The most recent method Geo2Vec~\cite{geo2vec} captures fine-grained geometric
information utilizing a signed distance field but fails to explicitly model detailed interactions along object boundaries.}
Existing representations still struggle to model polyline--polygon and
polygon--polygon relationships, which require local boundary behavior and global spatial context to be captured simultaneously.
Developing a unified representation for heterogeneous spatial objects raises
three technical challenges. First, the sampling process must capture narrow
boundary interactions without losing coverage of interiors and outer
space. Second, the same encoder must handle points, polylines, and polygons,
although these geometry types differ in dimensionality and interior structure.
Third, while canonical normalization is needed to compare geometric shape across
objects, it removes absolute size and location information that is required
for metric and range-based reasoning.

To address these challenges, we propose \textsc{SpatialVec}, a unified
representation learning framework for heterogeneous spatial objects.
SpatialVec first maps each object into a canonical geometry space while
retaining its original metadata and location features. It then applies a boundary-aware sampling strategy, allocating a larger share of the sampling budget to topology-sensitive boundary
regions of an object while retaining complementary samples from polygon interiors and the
outer-space. \textcolor{black}{At each sampled location, the framework evaluates three complementary geometric
fields: an SDF that measures the location's distance from the object, an
occupancy field that indicates whether it belongs to or lies near the object,
and a boundary field that identifies locations close to the object's boundary.} A \textcolor{black}{Transformer encoder processes these geometric observations and combines
global, boundary-focused, and distance-aware information into a compact spatial
representation of the object. During training, a reconstruction decoder uses
this representation to recover the geometric information observed at the
sampled locations, encouraging the learned representation to preserve both
overall shape and fine-grained boundary structure} to retain both broad geometric structure and local boundary evidence. The
learned code is finally combined with normalized metadata and original-location
encoding to form the SpatialVec embedding. SpatialVec achieves the best topological classification accuracy across all evaluated geometry pairs, improving the accuracy by up to 22.9\%. It also reduces the distance-estimation error by
$15.4 - 59.7\%$ across real-world geospatial
datasets. The largest gains are observed for the more challenging polyline--polygon and
polygon--polygon relationships, and these results indicate that preserving boundary-localized
evidence together with geometry and original spatial context is
particularly important for reasoning over heterogeneous spatial
objects.

\noindent
The main contributions of this work are as follows:

\noindent
\begin{itemize}
    \item We introduce a unified implicit-field representation for points, polylines, and polygons that combines signed distance, occupancy, and boundary
    observations within a common encoding framework.

    \item We develop a boundary-aware sampling strategy that
    concentrates observations near topology-sensitive boundaries while
    retaining coverage of polygon interiors and outer-space.

    \item We design a Transformer-based encoder with global, boundary-aware,
    and distance-aware aggregation, together with a self-supervised decoder
    that reconstructs the sampled spatial fields.

    \item We evaluate SpatialVec on two real-world geospatial datasets across
    topological relationship classification, distance estimation, and range-query
    prediction, demonstrating consistent improvements over existing spatial
    representation methods.
\end{itemize}

\section{Preliminaries}
\label{sec:preliminaries}

\noindent\textbf{Definition 1 (Spatial Object):}
A spatial object $E$ is a geographic entity represented in a two-dimensional
space. Its geometry is described by a sequence of $N$ point coordinates $\{p_1, p_2, ..., p_N\}$, where each point
$\mathbf{p}_i=(x_i,y_i)\in\mathbb{R}^{2}$.

\noindent\textbf{Definition 2 (Canonical Geometry Space):}
The canonical geometry space is a normalized coordinate system used to represent spatial objects independently of their original position and scale. Given the centroid $\mathbf{c}_E$ of an object \textcolor{black}{$E$} and the diagonal length $d_E$ of its bounding box, each coordinate \textcolor{black}{$\mathbf{p}_i$} is transformed as
\[
\widetilde{\mathbf{p}}_i
=
\frac{\mathbf{p}_i-\mathbf{c}_E}
{\max(d_E,\epsilon)},
\]
where $\epsilon$ is a small constant that prevents division by zero. This transformation centers the object at the origin and places geometries of different sizes in a common scale while preserving their relative shape and orientation.

\noindent\textbf{Definition 3 (Signed Distance Function (SDF) Field):}
\textcolor{black}{Given a canonical spatial object $\widetilde{E}$ and a query point $\mathbf{q}\in\mathbb{R}^{2}$, the SDF field $\phi_E(\mathbf{q})$ measures the distance from $\mathbf{q}$ to $\widetilde{E}$.}
If the object $\widetilde{E}$ is a polygon,
\[
\phi_E(\mathbf{q})
=
\begin{cases}
-d(\mathbf{q},\partial\widetilde{E}),
& \mathbf{q}\text{ lies inside }\widetilde{E},\\[2pt]
+d(\mathbf{q},\partial\widetilde{E}),
& \mathbf{q}\text{ lies outside }\widetilde{E},
\end{cases}
\]
\textcolor{black}{where $\partial\widetilde{E}$ denotes the set of points on the boundary $\widetilde{E}$ and
$d(\mathbf{q},\partial\widetilde{E})=\min_{\mathbf{p}\in \partial\widetilde{E}}\|\mathbf{q}-\mathbf{p}\|_2$
denotes the Euclidean distance from $\mathbf{q}$ to the closest point in
the set $\partial\widetilde{E}$. If the object $\widetilde{E}$ is a point or polyline, $\phi_E(\mathbf{q}) = +d(\mathbf{q},\partial\widetilde{E})$
since points and
polylines do not have interior regions.}

\noindent\textbf{Definition 4 (Occupancy and Boundary Fields):}
\label{def:occupancy_boundary}
The occupancy field shows whether a query point is part of, or very close to, a spatial object. For a polygon,
\[
o_E(\mathbf{q})=
\begin{cases}
1, & \mathbf{q}\text{ is inside }\widetilde{E},\\
0, & \mathbf{q}\text{ is outside }\widetilde{E}.
\end{cases}
\]
Since points and polylines do not have an interior, a query point is considered occupied when its distance from the object is smaller than a threshold $\tau_o$:
\[
o_E(\mathbf{q})=
\begin{cases}
1, & \phi_E(\mathbf{q})\leq\tau_o,\\
0, & \phi_E(\mathbf{q})>\tau_o.
\end{cases}
\]
The boundary field shows whether a query point lies on or near the object's boundary:
\[
b_E(\mathbf{q})=
\begin{cases}
1, & |\phi_E(\mathbf{q})|\leq\tau_b,\\
0, & |\phi_E(\mathbf{q})|>\tau_b,
\end{cases}
\]
where $\tau_o$ and $\tau_b$ are predefined distance thresholds.

\noindent\textbf{Definition 5 (Implicit Spatial Field):}
Given a canonical spatial object $\widetilde{E}$ and a query point
$\mathbf{q}\in\mathbb{R}^{2}$, the implicit spatial field is defined as
\[
\Psi_E(\mathbf{q})
=
\left[
\phi_E(\mathbf{q}),
o_E(\mathbf{q}),
b_E(\mathbf{q})
\right].
\]
It represents the geometry of $E$ through three field values evaluated at spatial query locations.

\noindent\textbf{Definition 6 (Spatial Embedding Learning):}
Given a collection of spatial objects $\mathcal{G}=\{E_j\}_{j=1}^{M}$,
spatial embedding learning aims to learn a mapping $F_{\theta}:E\rightarrow\mathbf{e}_E\in\mathbb{R}^{d}$,
where $\mathbf{e}_E$ is a fixed-dimensional representation of object $E$. In SpatialVec, the embedding is learned from sampled distance, occupancy, and boundary fields together with the object's spatial information. The same mapping is used for points, polylines, and polygons, allowing different geometry types to be represented in a common embedding space.
\section{Methodology}

\begin{figure*}[t]
    \centering
    \includegraphics[width=\textwidth]{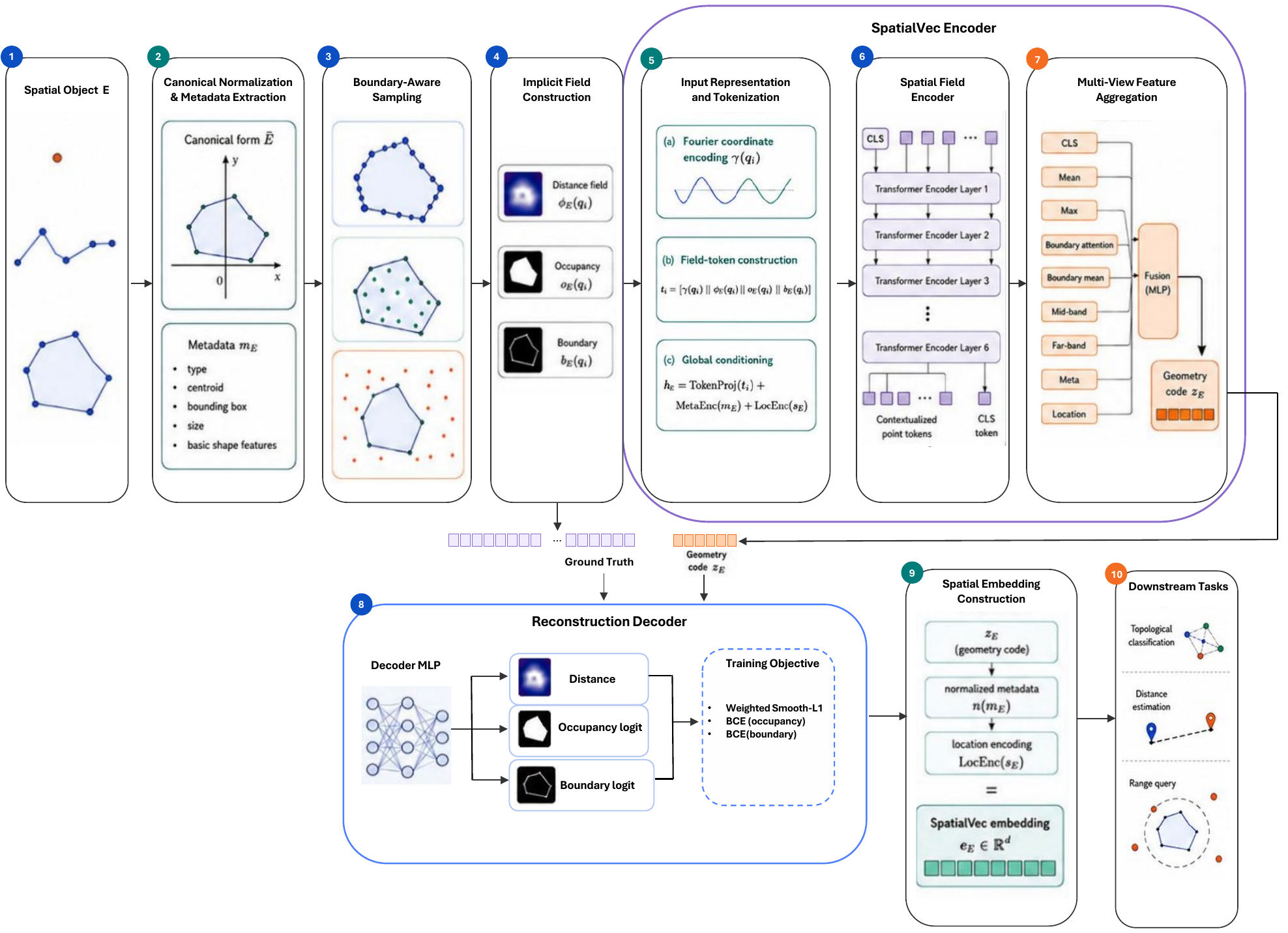}
    \vspace{-4pt}
    \caption{End-to-end data flow and architecture of SpatialVec framework.}
    \label{fig:spatialvec_overview}
    \vspace{-8pt}
\end{figure*}

Given a spatial object $E$, SpatialVec first maps it to the canonical space
while retaining metadata and original-location information. \textcolor{black}{After the mapping, SpatialVec applies \textit{Adaptive Boundary-Aware Sampling}, where
the number and distribution of sampled points are determined by the geometric
characteristics of each object, with greater emphasis on boundary regions while
retaining coverage of polygon interiors and the surrounding space.} After the sampling, \textit{SpatialVec Encoder}
encodes each sampled point to a \textcolor{black}{geometry code $\mathbf{z}_E$}, \textcolor{black}{which is a compact
representation of the spatial object that summarizes the information captured
from the sampled points, including its overall geometry, boundary structure,
and spatial characteristics.} During training, the \textit{Reconstruction Decoder} uses the encoded representation to
recover the spatial information associated with the sampled points. Finally,
$\mathbf{z}_E$ is combined with metadata and location features to form the
SpatialVec embedding $\mathbf{e}_E$. Fig.~\ref{fig:spatialvec_overview}
illustrates the end-to-end data flow and architecture of SpatialVec.

As mentioned earlier in this section, before sampling, each spatial object $E$ is transformed into its canonical form $\widetilde{E}$. \textcolor{black}{During this mapping,} SpatialVec also retains a 17-dimensional metadata vector $\mathbf{m}_E$ computed from the original geometry. Here, metadata refers to object-level attributes that describe what type of geometry the object is, where it is located, how large it is, and its overall shape. Specifically, $\mathbf{m}_E$ contains three geometry-type indicators, six centroid and bounding-box coordinates, five size-related features comprising width, height, bounding-box diagonal, log-transformed area, and log-transformed length, and three shape descriptors: aspect ratio, fill ratio, and compactness. The rest of this section describes each component of the our methodology in detail.

\subsection{Adaptive Boundary-Aware Sampling} 
\textcolor{black}{SpatialVec samples query points directly in the canonical geometry space.
Because each object has already been centered and scaled, geometric properties
used by the sampler are measured in the same normalized coordinate system.
Rather than assigning the same number of query points to every object,
SpatialVec determines an object-specific sampling budget $N_E$ from the
geometry of $\widetilde{E}$. Objects with longer boundaries receive additional
boundary observations, while polygons with larger interiors receive additional
interior observations. The value of $N_E$ therefore varies across objects,
while $N_{\min}$ and $N_{\max}$ only specify the allowable lower and upper
limits of the sampling budget. The resulting object-specific budget is then
distributed among the boundary region, polygon interior, and outer-space.}
The sampling budget is adapted to the geometry of each object. Let $\ell_E$ denote the canonical boundary length for polygons, the
canonical line length for polylines, and $0$ for points. Also, let $A_E$ denote the
canonical polygon area and is set to $0$ for non-polygon geometries, while
$\delta_E=1$ for polygons and $0$ otherwise. For a canonical
object $\widetilde{E}$, we define

\vspace{-10pt}
\begin{equation}
\begin{aligned}
N_E =
\min\Big\{N_{\max},
\max\Big\{N_{\min},
\Big\lceil N_0(&1
+\lambda_{\ell}\ell_E\\
+\lambda_a\delta_E\sqrt{A_E})\Big\rceil
\Big\}\Big\}.
\end{aligned}
\label{eq:adaptive_budget}
\end{equation}
Let $W_{\mathrm{bnd}}$ and $W_{\mathrm{int}}$ denote the sampling weights assigned to boundary and interior, respectively. The value of $W_{\mathrm{int}}$ is set to $0$ if the canonical object $\widetilde{E}$ is a polygon. The numbers of samples for boundary ($N_{\mathrm{bnd}}$), interior ($N_{\mathrm{int}}$), and outer-space region ($N_{\mathrm{space}}$) are calculated as: 

\begin{equation}
\begin{aligned}
N_{\mathrm{bnd}} &=
\left\lfloor W_{\mathrm{bnd}} N_E\right\rfloor, \qquad
N_{\mathrm{int}} =
\left\lfloor W_{\mathrm{int}} N_E\right\rfloor,\\
N_{\mathrm{space}} &= N_E-N_{\mathrm{bnd}}-N_{\mathrm{int}},
\end{aligned}
\label{eq:adaptive_allocation}
\end{equation}

\begin{algorithm}[t]
\caption{Boundary-Aware Sampling}
\label{alg:spatialvec_sampling}

\KwObjective{
Generate an object-adaptive set of query points with increased coverage near
geometrically informative regions
}

\KwInput{
Canonical spatial object $\widetilde{E}$;
minimum and maximum budgets $N_{\min},N_{\max}$
}

\KwOutput{
Sampled implicit-field representation $\mathcal{S}_E$
}

$\ell_E \gets$ Canonical length of $\widetilde{E}$\;

\eIf{$\widetilde{E}$ is a polygon}{
    $A_E \gets$ area of $\widetilde{E}$\;
    $\delta_E \gets 1$\;
}{
    $A_E \gets 0$\;
    $\delta_E \gets 0$\;
}

$N_E
\gets
\min\!\left(
N_{\max},
\max\!\left(
N_{\min},
\left\lceil
N_0
\left(
1+\lambda_{\ell}\ell_E
+\lambda_a\delta_E\sqrt{A_E}
\right)
\right\rceil
\right)
\right)$\;

$w_{\mathrm{bnd}}
\gets
1+\lambda_b\ell_E$\;

$w_{\mathrm{int}}
\gets
\delta_E\left(1+\lambda_i\sqrt{A_E}\right)$\;

$w_{\mathrm{space}}
\gets 1$\;

$W
\gets
w_{\mathrm{bnd}}
+w_{\mathrm{int}}
+w_{\mathrm{space}}$\;

$N_{\mathrm{bnd}}
\gets
\left\lfloor
N_E\frac{w_{\mathrm{bnd}}}{W}
\right\rfloor$\;

$N_{\mathrm{int}}
\gets
\left\lfloor
N_E\frac{w_{\mathrm{int}}}{W}
\right\rfloor$\;

$N_{\mathrm{space}}
\gets
N_E-N_{\mathrm{bnd}}-N_{\mathrm{int}}$\;

$\mathcal{Q}_{\mathrm{bnd}}
\gets
\operatorname{SampleBoundary}
(\widetilde{E},N_{\mathrm{bnd}})$\;

\eIf{$\widetilde{E}$ is a polygon}{
    $\mathcal{Q}_{\mathrm{int}}
    \gets
    \operatorname{SampleInterior}
    (\widetilde{E},N_{\mathrm{int}})$\;
}{
    $\mathcal{Q}_{\mathrm{int}}
    \gets
    \emptyset$\;
}

$\mathcal{Q}_{\mathrm{space}}
\gets
\operatorname{SampleOuterSpace}
(N_{\mathrm{space}})$\;

$\mathcal{Q}_E
\gets
\mathcal{Q}_{\mathrm{bnd}}
\cup
\mathcal{Q}_{\mathrm{int}}
\cup
\mathcal{Q}_{\mathrm{space}}$\;

$\mathcal{Q}_E
\gets
\operatorname{Shuffle}(\mathcal{Q}_E)$\;

\textbf{return} $\mathcal{Q}_E$\;

\end{algorithm}




\noindent
Algorithm~\ref{alg:spatialvec_sampling} summarizes the boundary-aware
sampling procedure. First, SpatialVec computes the geometric properties of the
canonical object $\widetilde{E}$, including its canonical length and polygon
area when applicable (lines~1--7). These properties are used to determine the
object-specific sampling budget $N_E$ according to
Eq.~\eqref{eq:adaptive_budget} (line~8). The total budget is then distributed among the boundary, interior, and
outer-space regions. The corresponding sampling weights are computed in
lines~9--12, and the resulting values of $N_{\mathrm{bnd}}$,
$N_{\mathrm{int}}$, and $N_{\mathrm{space}}$ are obtained in
lines~13--15 according to Eq.~\eqref{eq:adaptive_allocation}. The function $\operatorname{SampleBoundary}(\widetilde{E},N_{\mathrm{bnd}})$ in
line~16 samples $N_{\mathrm{bnd}}$ points from the boundary region of the
object $\widetilde{E}$. This sampling procedure is detailed in
Section~\ref{subsec:boundary-sampling}. For polygon geometries,
$\operatorname{SampleInterior}(\widetilde{E},N_{\mathrm{int}})$ in
lines~17--20 samples $N_{\mathrm{int}}$ points from the polygon interior,
whereas points and polylines receive no interior samples. Similarly the function
$\operatorname{SampleOuterSpace}(N_{\mathrm{space}})$ in line~21 samples
$N_{\mathrm{space}}$ points from the surrounding canonical space.
The functions $\operatorname{SampleInterior}$ and $\operatorname{SampleOuterSpace}$ are detailed in Section \ref{subsec:interior-sample}.
The boundary, interior, and outer-space samples are then combined to form
$\mathcal{Q}_E$ (line~22) and randomly shuffled (line~23).

\subsubsection{Boundary Sampling}
\label{subsec:boundary-sampling}
The number of boundary samples $N_{\mathrm{bnd}}$ is determined by the
adaptive allocation in Eq.~\eqref{eq:adaptive_allocation}. For a
polygon or polyline, a boundary segment is selected according to its length.
If the canonical boundary contains segments with lengths
$\ell_1,\ldots,\ell_K$, segment $k$ is selected with probability

\vspace{-8pt}
\begin{equation}
    \Pr(k)=\frac{\ell_k}{\sum_{j=1}^{K}\ell_j}.
\end{equation}
This gives longer segments proportionally more samples and avoids making the
sampling distribution depend on the number of vertices. After selecting a
segment, points are sampled uniformly along it, ensuring that its distance from the line segment is $\leq \tau_b$. The samll offset between two consecutive samples is defined as:

\vspace{-10pt}
\begin{equation}
    \mathbf{T}
    =
    (1-f)\widetilde{\mathbf{x}}_k
    +
    f\widetilde{\mathbf{x}}_{k+1}
    +
    r
    \begin{bmatrix}
        \cos\theta\\
        \sin\theta
    \end{bmatrix},
    \label{eq:boundary_sampling}
\end{equation}
where $f\sim\mathcal{U}(0,1)$ determines the position on the segment,
$\theta\sim\mathcal{U}(0,2\pi)$ determines the offset direction, and
$r\in\{0,0.004,0.012,0.03,0.075\}$ specifies the offset distance.
The value $r=0$ produces samples directly on the boundary, while the remaining
values create several outer layers at increasing distances. This
provides dense observations of the region in which the implicit field changes
most rapidly. For a point geometry, samples are instead drawn around the
canonical origin using radii
$\{0.004,0.012,0.03,0.075,0.18\}$.

\subsubsection{Interior and Outer-Space Sampling}
\label{subsec:interior-sample}
\textcolor{black}{
In the case of interior sampling, points and polylines receive no interior samples because they do not have a two-dimensional interior region.}
For polygons, candidate points are sampled independently and uniformly within the minimum and maximum coordinate ranges of the canonical bounding box. Each candidate point is then evaluated using a point-in-polygon test, and retained only if it is located inside the polygon, and distance from the polygon boundary is $> \tau_b$. This process is repeated until the required number of interior samples ($N_{\mathrm{int}}$) is obtained.

\textcolor{black}{For outer-space sampling, candidate points are sampled independently and uniformly over the corresponding horizontal and vertical extents of the surrounding canonical space. Thus, each location within this sampling region has an equal probability of being selected. If the object is a polygon, point-in-polygon test ensures that the candidate point is outside the polygon. Similarly to the interior sampling, a candidate point is retained only if the distance from the object boundary is $> \tau_b$. The sampling is repeated until the required number of outer-space samples ($N_{\mathrm{space}}$) is obtained.
}






\subsection{Unified Implicit-Field Representation}

\textcolor{black}{After boundary-aware sampling, SpatialVec represents each sampled location using
multiple complementary descriptions of the underlying geometry.} Rather than
relying only on the SDF value, SpatialVec jointly uses the three fields defined
in Section~\ref{sec:preliminaries}: the SDF field, occupancy field, and boundary
field. 

Each field captures a different aspect of the spatial object. The SDF field
describes how far the sampled point is from the object and whether the point is inside or outside the object in case of a polygon, providing continuous
information about its geometric structure. The occupancy field provides
information about whether the sampled location belongs to or lies close to the
object, helping characterize its spatial extent. The boundary field explicitly
identifies locations near the object boundary, where small geometric changes
are particularly important for distinguishing spatial relationships. By
combining these three complementary sources of information, SpatialVec captures
both the overall structure of an object and fine-grained information around its
boundary. Given the sampled locations
$\mathcal{Q}_E=\{\mathbf{q}_i\}_{i=1}^{N_E}$, the resulting unified
implicit-field representation is

\begin{equation}
\mathcal{S}_E
=
\left\{
\left(
\mathbf{q}_i,
\phi_E(\mathbf{q}_i),
o_E(\mathbf{q}_i),
b_E(\mathbf{q}_i)
\right)
\right\}_{i=1}^{N_E}.
\label{eq:implicit_field_representation}
\end{equation}
This representation provides the same input form for points, polylines, and
polygons despite their different geometric structures. Therefore, SpatialVec
does not require a separate representation scheme for each geometry type.
The unified implicit-field representation $\mathcal{S}_E$ is subsequently
converted into input tokens and processed by the SpatialVec encoder.

\subsection{SpatialVec Encoder}
The SpatialVec encoder converts the sampled implicit-field observations through three main stages. First,
\textit{Input Representation and Tokenization} combines Fourier-encoded
canonical coordinates with the corresponding distance, occupancy, boundary,
and distance-band values, together with metadata and original-location
information.
Next, \textit{Transformer-Based Spatial Field Encoding} processes the
query-point tokens using self-attention, while a learnable CLS token aggregates
information across the sampled locations. Finally, \textit{Multi-View Feature
Aggregation} summarizes the contextualized representations using global,
boundary-aware, and distance-aware views. These complementary views are fused
to produce the 256-dimensional \textcolor{black}{geometry code $\mathbf{z}_E$} which compactly summarizes the spatial characteristics of
the object.

\subsubsection{Input Representation and Tokenization}

The sampled query points and their associated field values are converted into
a sequence of input tokens for the encoder. Each token represents one query
point and combines its canonical location with the corresponding distance,
occupancy, and boundary information.

\noindent
\textbf{\textit{Multi-frequency Coordinate Encoding.}}
\textcolor{black}{To capture spatial patterns at different scales, the canonical coordinates of
each sampled point are represented using multiple Fourier frequencies. Let
$\mathbf{q}_i=(x_i^q,y_i^q)$ denote the canonical coordinates of sampled
point $i$, and let
$F=\{1,2,4,8,16,32,64,128\}$ denote the set of eight frequency levels.
The coordinate encoding is defined as}
\begin{equation}
\gamma(\mathbf{q}_i)
=
\left[
\mathbf{q}_i
\;\Vert\;
\big\{
\sin(\pi f\mathbf{q}_i)
\;\Vert\;
\cos(\pi f\mathbf{q}_i)
\big\}_{f\in F}
\right].
\label{eq:coordinate_encoding}
\end{equation}
For each $f\in F$, the frequency $f$ is multiplied by both coordinates of $\mathbf{q}_i$, giving $f\mathbf{q}_i=(f\cdot x_i^q,f\cdot y_i^q)$, and the sine and cosine functions are applied separately to both coordinates. Lower frequencies capture broader
spatial variations, while higher frequencies capture finer geometric
variations. Together with the original two-dimensional coordinates, these
features form the 34-dimensional coordinate representation $\gamma(\mathbf{q}_i)$.

\noindent
\textbf{\textit{Field-token Construction.}}
The coordinate encoding is combined with the previously computed implicit field
values. Each query point is also assigned to one of three distance bands:
near, middle, or far. The token for query point $\mathbf{q}_i$ is constructed as
\begin{equation}
\mathbf{t}_i =
\left[
\gamma(\mathbf{q}_i)
\Vert \phi_E(\mathbf{q}_i)
\Vert o_E(\mathbf{q}_i)
\Vert b_E(\mathbf{q}_i)
\Vert \operatorname{band}\big(\phi_E(\mathbf{q}_i)\big)
\right].
\label{eq:field_token}
\end{equation}
\textcolor{black}{Here, $\operatorname{band}(\phi_E(\mathbf{q}_i))$ is a three-dimensional
one-hot vector that indicates whether the sampled point lies in the near,
middle, or far distance band, using thresholds of $0.06$ and $0.22$.}
The resulting token is $\mathbf{t}_i\in\mathbb{R}^{40}$. Each 40-dimensional token is then projected to the model
dimension of 256 using a linear layer, GELU activation, and layer normalization.
The field values are included in the encoder input so that each token contains
both the spatial location of the sampled point and the geometric information
observed at that location.

\noindent
\textbf{\textit{Metadata and Location Integration.}}
\textcolor{black}{Object-level metadata and original-location information are added to the
256-dimensional representation of each field token before Transformer encoding.} In parallel, the centroid and bounding-box features are transformed using a
multi-frequency location encoding and projected through a separate MLP. These two global representations are added to every point token:
\begin{equation}
\mathbf{h}_i
=
\operatorname{TokenProj}(\mathbf{t}_i)
+
\operatorname{MetaEnc}(\mathbf{m}_E)
+
\operatorname{LocEnc}(\mathbf{s}_E).
\label{eq:conditioned_token}
\end{equation}
Here, $\operatorname{TokenProj}$ is a linear projection followed by GELU
activation and layer normalization that maps
$\mathbf{t}_i\in\mathbb{R}^{40}$ to a 256-dimensional vector.
$\operatorname{MetaEnc}$ applies the same sequence of operations to the
17-dimensional metadata vector. $\operatorname{LocEnc}$ first applies
multi-frequency Fourier encoding to the six original-location features and
then maps them to 256 dimensions using a two-layer MLP. Their sum,
$\mathbf{h}_i\in\mathbb{R}^{256}$, is the point representation passed to the
Transformer encoder. This allows every token to retain its local field information while also
receiving information about the geometry type, size, and original spatial
location.

\noindent
\textbf{\textit{Set-based Representation.}}
The sampled query points form an unordered set. Therefore, no sequential
position encoding is added to the token sequence. The location of each point
is already represented by its Fourier-encoded coordinates. Consequently,
changing the order of the sampled points does not change the geometry being
represented. Finally, a learnable classification token is prepended to the
point-token sequence before it is passed to the Transformer encoder.


\subsubsection{Transformer-Based Spatial Field Encoding}

A learnable CLS token is prepended to the point representations
$\mathbf{h}_1,\ldots,\mathbf{h}_{N_E}$, and the resulting sequence is processed
by a six-layer Transformer encoder. Each layer uses a hidden dimension of 256,
eight attention heads, a feed-forward dimension of 1,024, GELU activation,
pre-normalization, and a dropout rate of 0.1.

In each layer, self-attention updates every point token using information from
the CLS token and all other point tokens. Therefore, each point representation
retains its local field information while also incorporating the spatial
context of the remaining sampled locations. The CLS token is updated in the same way and aggregates information from all point tokens. After six layers,
the encoder outputs contextualized point representations and a contextualized
CLS representation, which are used for Multi-View Feature Aggregation.


\subsubsection{Multi-View Feature Aggregation}

\textcolor{black}{Although the number of sampled points $N_E$ varies across objects, the
aggregation operations convert these variable-size token sets into
fixed-dimensional representations.}
SpatialVec converts these outputs into fixed-size summaries using nine views:
the CLS vector, mean and max pooling over all point vectors, attention-based
boundary pooling, mean pooling over transformed boundary vectors, mean pooling
over the middle- and far-distance bands, the metadata vector, and the
original-location vector.
These nine vectors are concatenated and passed through a fusion MLP:

\vspace{-8pt}
\begin{equation}
\begin{aligned}
\mathbf{u}_E = \big[
&\mathbf{v}_{\mathrm{cls}}
\Vert \mathbf{v}_{\mathrm{mean}}
\Vert \mathbf{v}_{\mathrm{max}}
\Vert \mathbf{v}_{\mathrm{battn}} \\
&\Vert \mathbf{v}_{\mathrm{bnd}}
\Vert \mathbf{v}_{\mathrm{mid}}
\Vert \mathbf{v}_{\mathrm{far}}
\Vert \mathbf{v}_{\mathrm{meta}}
\Vert \mathbf{v}_{\mathrm{loc}}
\big],
\end{aligned}
\label{eq:multiview_concat}
\end{equation}

\begin{equation}
\mathbf{z}_E=\operatorname{Fuse}(\mathbf{u}_E)
\in\mathbb{R}^{256}.
\label{eq:geometry_code}
\end{equation}
Here, $\mathbf{v}_{\mathrm{cls}}$ is the contextualized CLS representation,
$\mathbf{v}_{\mathrm{mean}}$ and $\mathbf{v}_{\mathrm{max}}$ are the mean- and
max-pooled point representations, and $\mathbf{v}_{\mathrm{battn}}$ is the
attention-weighted boundary representation. The vector
$\mathbf{v}_{\mathrm{bnd}}$ is obtained by mean pooling the transformed
boundary-point representations, while $\mathbf{v}_{\mathrm{mid}}$ and
$\mathbf{v}_{\mathrm{far}}$ summarize the middle- and far-distance bands,
respectively. Finally, $\mathbf{v}_{\mathrm{meta}}$ and
$\mathbf{v}_{\mathrm{loc}}$ denote the encoded metadata and original-location
representations. Their concatenation forms
$\mathbf{u}_E\in\mathbb{R}^{2304}$, and $\operatorname{Fuse}$ denotes the
fusion MLP that maps $\mathbf{u}_E$ to the geometry code
$\mathbf{z}_E\in\mathbb{R}^{256}$.

The resulting geometry code contains object-level, boundary, distance-band,
metadata, and location information.


\subsection{Reconstruction Decoder}
The reconstruction decoder uses the geometry code and the contextualized representation of each query point to recover the sampled spatial fields. For every query point, its encoder representation is concatenated with the global geometry code and passed through a two-layer MLP. Three prediction heads then estimate the implicit distance, occupancy, and boundary values. By reconstructing these fields at all sampled locations, the decoder provides the self-supervised training signal for the encoder. The distance loss is weighted more heavily near the boundary, while occupancy and boundary prediction are trained using binary cross-entropy. Together, these objectives encourage the geometry code to preserve both the overall spatial structure and the fine boundary details of the object.

\subsubsection{Implicit Field Reconstruction and Training Objective}



Let $\mathbf{p}_i\in\mathbb{R}^{256}$ denote the contextualized encoder
representation of query point $\mathbf{q}_i$. The geometry code
$\mathbf{z}_E$ is concatenated with each point representation and processed
by a two-layer reconstruction MLP, giving
$\mathbf{d}_i=\operatorname{Dec}(\mathbf{p}_i\Vert\mathbf{z}_E)$.
Three linear heads use $\mathbf{d}_i$ to predict the implicit distance
$\hat{\phi}_i$, occupancy logit $a_i^{o}$, and boundary logit $a_i^{b}$.
The occupancy and boundary probabilities are obtained by applying the sigmoid
function to their logits. To emphasize samples near the geometry boundary, the distance loss assigns
the following weight to query point $\mathbf{q}_i$:

\vspace{-8pt}
\begin{equation}
w_i
=
\min\left(
\frac{1}{|\phi_E(\mathbf{q}_i)|+0.02},
10
\right).
\label{eq:sdf_weight}
\end{equation}
When the inverse-distance weight becomes greater than $10$, a small number of
samples extremely close to the boundary can contribute disproportionately to
the loss and dominate the optimization, reducing the influence of the remaining
samples. We therefore set the upper bound to $10$, which limits the contribution
of any individual sample to at most ten times the base weight. This value
provides strong emphasis to near-boundary regions while keeping their influence
within one order of magnitude and maintaining a balanced training objective.
\textcolor{black}{The SDF field prediction is trained using weighted Smooth-L1 loss, while occupancy and
boundary prediction use binary cross-entropy with logits}:


\begin{equation}
\begin{aligned}
\mathcal{L}_{\mathrm{sdf}}
&=
\frac{
\sum_{i=1}^{N}
w_i\,\operatorname{SmoothL1}
\left(\hat{\phi}_i,\phi_E(\mathbf{q}_i)\right)}
{\sum_{i=1}^{N}w_i},
\\
\mathcal{L}_{\mathrm{occ}}
&=
\frac{1}{N}\sum_{i=1}^{N}
\operatorname{BCE}
\left(a_i^{o},o_E(\mathbf{q}_i)\right),
\\
\mathcal{L}_{\mathrm{bnd}}
&=
\frac{1}{N}\sum_{i=1}^{N}
\operatorname{BCE}
\left(a_i^{b},b_E(\mathbf{q}_i)\right).
\end{aligned}
\label{eq:reconstruction_losses}
\end{equation}

\noindent
\textcolor{black}{Here, $\mathcal{L}{\mathrm{sdf}}$, $\mathcal{L}{\mathrm{occ}}$, and $\mathcal{L}_{\mathrm{bnd}}$ denote the signed distance field, occupancy, and boundary reconstruction losses, respectively.}

A small $L_2$ penalty on $\mathbf{z}_E$ limits the magnitude of the geometry
code. The total training objective is

\begin{equation}
\begin{aligned}
\mathcal{L}
={}&
\mathcal{L}_{\mathrm{sdf}}
+0.55\,\mathcal{L}_{\mathrm{occ}}
&+0.70\,\mathcal{L}_{\mathrm{bnd}}
+10^{-4}\|\mathbf{z}_E\|_2^2.
\end{aligned}
\label{eq:total_loss}
\end{equation}

\subsection{Spatial Embedding Construction}

After \textcolor{black}{encoding}, SpatialVec forms the final embedding $\mathbf{e}_E$ by concatenating the
learned geometry code with the normalized metadata and the Fourier encoding of
the original centroid and bounding-box features:

\begin{equation}
\mathbf{e}_E
=
\left[
\mathbf{z}_E
\;\Vert\;
\widetilde{\mathbf{m}}_E
\;\Vert\;
\gamma_{\mathrm{loc}}(\mathbf{s}_E)
\right]
\in\mathbb{R}^{477},
\label{eq:spatialvec_embedding}
\end{equation}
where $\mathbf{z}_E\in\mathbb{R}^{256}$ is the geometry code,
$\widetilde{\mathbf{m}}_E\in\mathbb{R}^{17}$ is the normalized metadata
vector, and $\gamma_{\mathrm{loc}}(\mathbf{s}_E)\in\mathbb{R}^{204}$ is the
Fourier encoding of the six centroid and bounding-box features. Thus, the
final embedding $\mathbf{e}_E$ has $256+17+204=477$ dimensions. The geometry code represents the sampled implicit field, while the metadata
and location components retain the geometry type, size, and original spatial
position.


\section{Experimental Evaluation}


We evaluate SpatialVec by addressing the following research questions. \textbf{RQ1:} How effectively do SpatialVec embeddings preserve geometric
relationships, relative spatial distances, and spatial location and distribution
information? \textbf{RQ2:} What is the computational efficiency of SpatialVec's
embedding-generation process compared with existing methods? \textbf{RQ3:} How do the key components of SpatialVec contribute to the
quality of the learned spatial embeddings?

\vspace{2pt}
\noindent
\textbf{Training Configuration.} \textcolor{black}{The embedding model is trained for 30 epochs using the AdamW optimizer with a learning rate of $3 \times 10^{-4}$, weight decay of $10^{-4}$, and a batch size of 32.}


\subsection{Datasets}
We use two real-world geospatial datasets from the recent work~\cite{poly2vec}, NYC and Singapore, both derived
from OpenStreetMap~\cite{appli_urban}. Each dataset contains point, polyline, and
polygon geometries and is used to evaluate whether the learned embeddings
capture geometric and spatial information effectively in downstream GeoAI
tasks.

\noindent
\textbf{NYC.}
The NYC dataset covers New York City and contains 14,943 points of interest,
139,512 road geometries, and 1,153,008 building geometries.

\noindent
\textbf{Singapore.}
The Singapore dataset contains 4,347 points of interest, 45,634 road
geometries, and 109,877 building geometries from the Singapore region.

\subsection{Baselines}
We compare SpatialVec with the following spatial representation learning methods:

\noindent
\textbf{THEORY.}
THEORY~\cite{theory} learns spatial embedding by encoding the spatial distribution
and relationships of geo-entities.

\noindent
\textbf{Hist2Vec.}
Hist2Vec~\cite{hist2vec} represents spatial data using multiresolution histogram features and
learns fixed-dimensional embeddings through an unsupervised autoencoder.

\noindent
\textbf{Poly2Vec.}
Poly2Vec~\cite{poly2vec} provides a unified representation for point, polyline, and polygon
geometries using a Fourier-based polymorphic encoder. 

\noindent
\textbf{Geo2Vec.}
Geo2Vec~\cite{geo2vec} represents heterogeneous spatial objects through sampled implicit
SDF field and learns fixed-dimensional embeddings that capture shape,
location, and spatial relationships.

We also include several geometry-specific encoders as baselines. For point objects, DIRECT \cite{direct}
uses the original coordinates directly, TILE \cite{tile} partitions the spatial domain
into discrete cells and represents each location through a tile embedding,
WARP \cite{warp} applies a coordinate-wrapping mechanism, \textcolor{black}{and GRID~\cite{nuftspec} uses multi-scale positional encoding to represent spatial locations.}



\begin{table*}[t]
\centering
\caption{Model accuracy on topological relationship classification.
All values are scaled by $\times 10^{-2}$.}
\label{tab:topological-classification}

\small
\setlength{\tabcolsep}{4pt}
\renewcommand{\arraystretch}{1.05}

\begin{tabular}{@{}lccccc ccccc@{}}
\toprule
& \multicolumn{5}{c}{Singapore}
& \multicolumn{5}{c}{NYC} \\
\cmidrule(lr){2-6}
\cmidrule(lr){7-11}

Method
& Pt--Pl$\uparrow$
& Pt--Pg$\uparrow$
& Pl--Pl$\uparrow$
& Pl--Pg$\uparrow$
& Pg--Pg$\uparrow$
& Pt--Pl$\uparrow$
& Pt--Pg$\uparrow$
& Pl--Pl$\uparrow$
& Pl--Pg$\uparrow$
& Pg--Pg$\uparrow$ \\
\midrule




DIRECT
& 82 & 84 & 73 & 36 & 35
& 84 & 90 & 74 & 49 & 44 \\

TILE
& 79 & 70 & 50 & 45 & 41
& 65 & 78 & 50 & 49 & 40 \\

WARP
& 88 & 88 & 71 & 47 & 47
& 88 & 88 & 73 & 55 & 38 \\

GRID
& 84 & 84 & 69 & 45 & 45
& 82 & 89 & 73 & 51 & 38 \\

THEORY
& 89 & 90 & 71 & 45 & 45
& 89 & 90 & 73 & 59 & 45 \\

Hist2Vec
& 89 & 62 & 94 & 44 & 47
& 91 & 62 & \underline{94} & 44 & 47 \\

Poly2Vec
& \underline{95} & 94 & 81 & 50 & 70
& 95 & \underline{98} & 83 & 64 & 68 \\

Geo2Vec
& \textbf{98} & \underline{96} & \underline{96}
& \underline{61} & \underline{75}
& \underline{98} & \textbf{99} & \textbf{98}
& \underline{67} & \underline{70} \\

\midrule

\textbf{SpatialVec}
& \textbf{98} & \textbf{98} & \textbf{97}
& \textbf{64} & \textbf{80}
& \textbf{99} & \textbf{99} & \textbf{98}
& \textbf{80} & \textbf{86} \\

\bottomrule
\end{tabular}
\vspace{-8pt}
\end{table*}

\subsubsection{Topological Relationship Classification}

In this evaluation, a downstream classifier receives the embeddings of two spatial objects and
predicts their topological relationship such as disjoint, touching, overlap, containment, and intersection.
We use Pt, Pl, and Pg to denote point, polyline, and polygon geometries, respectively. Following the established evaluation protocol, we consider five geometry-pair
settings: Pt--Pl, Pt--Pg, Pl--Pl, Pl--Pg, and Pg--Pg. The Pt--Pl, Pt--Pg,
and Pl--Pl settings are formulated as binary classification tasks, whereas
Pl--Pg and Pg--Pg contain multiple topological relationship classes. All
embedding methods are evaluated using the same training and test samples and
the same downstream classifier configuration. We report the topological classification
accuracy results in
Table~\ref{tab:topological-classification}.


\subsection{Effectiveness in Spatial Reasoning Tasks}

To address RQ1, we evaluate the quality of the learned embeddings through three spatial
reasoning tasks: topological relationship classification, distance estimation,
and range-query's selectivity prediction.

For topological relationship classification and distance estimation, we follow
the evaluation protocol established by prior spatial representation learning
methods, including Poly2Vec and Geo2Vec. Specifically, we use the same NYC and
Singapore datasets, the same geometry-pair categories, and the same downstream
task formulations to ensure a controlled and directly comparable evaluation.
Only the input object representations are replaced by the embeddings generated
by each evaluated method. As introduced in Definition~\ref{def:occupancy_boundary}, the occupancy and
boundary thresholds are set to $\tau_o=\tau_b=0.025$ in the canonical
coordinate space and are kept fixed for all experiments. \textcolor{black}{Values of parameters $N_{min}$ and $N_{max}$ are set to $512$ and $2048$, respectively.} 




Table~\ref{tab:topological-classification} shows that SpatialVec achieves the
best or tied-best accuracy across all ten settings. The gains are modest for
the simpler binary tasks, where several methods already approach ceiling
performance, but become clearer for the more difficult Pl--Pg and Pg--Pg
settings. Relative to the strongest baseline, SpatialVec improves Pl--Pg
accuracy by $4.9\%$ in Singapore and \textcolor{black}{$19.4\%$} in NYC. For Pg--Pg, the relative
gains increase to $6.7\%$ and $22.9\%$, respectively, showing a larger advantage
when the relationship depends on detailed interactions between extended
geometries. These results suggest that SpatialVec retains finer local geometric information
than the competing representations. THEORY mainly captures spatial position,
while Hist2Vec aggregates geometry into histogram statistics, which can weaken
local boundary details. Poly2Vec preserves global shape through frequency-domain
features, but close boundary interactions may be less explicit. Geo2Vec performs
better than most baselines because it uses an implicit SDF field. Geo2Vec performs better than most baselines because it uses an implicit distance
field. SpatialVec performs the best by coupling boundary-aware sampling with an
encoder-decoder objective that compresses distance, occupancy, and boundary
fields into a shared \textcolor{black}{spatial representation} and reconstructs them during \textcolor{black}{encoding}.

\subsubsection{Distance Estimation}

A regression model receives the embeddings of two spatial objects and predicts
their ground-truth spatial distance. Because the prediction is made without
directly accessing their coordinates, a lower estimation error indicates that
the embeddings preserve the relative spatial distance between the objects.
We follow the same dataset construction and geometry-pair definitions used in
the prior evaluation protocol. The experiments include Pt--Pl, Pt--Pg,
Pl--Pl, Pl--Pg, and Pg--Pg pairs, and all representation methods are evaluated
using identical training and test splits and the same regression model. Several baselines \cite{t2vec,tjepa,nuftspec} in Table \ref{tab:topological-classification} are designed for a single geometry family, so
applying them would require combining different encoders
or changing the original method. We omit such hybrid variants \cite{direct,tile,warp} to keep the
distance comparison consistent with the published setups. Mean
absolute error (MAE) is used as the evaluation metric, where a lower value
indicates better preservation of distance information. The results are
reported in Table~\ref{tab:distance_estimation}.

\begin{table}[t]
\centering
\caption{Comparing MAE on distance estimation.
All MAE values are scaled by $\times 10^{-3}$.}
\label{tab:distance_estimation}

\scriptsize
\setlength{\tabcolsep}{6pt}
\renewcommand{\arraystretch}{1.05}

\begin{tabular}{@{}lccc ccc@{}}
\toprule
& \multicolumn{3}{c}{Singapore}
& \multicolumn{3}{c}{NYC} \\
\cmidrule(lr){2-4}
\cmidrule(lr){5-7}

Method
& Pg--Pt$\downarrow$
& Pl--Pg$\downarrow$
& Pg--Pg$\downarrow$
& Pg--Pt$\downarrow$
& Pl--Pg$\downarrow$
& Pg--Pg$\downarrow$ \\
\midrule

TILE 
& 99.9 & 115.5 & 114.3
& 127.6 & 154.3 & 167.4 \\

THEORY
& 24.30 & 25.00 & 25.00
& 26.20 & 26.60 & 27.40 \\

Hist2Vec
& 16.20 & 22.47 & 30.31
& 21.80 & 27.42 & 32.44 \\

Poly2Vec
& 13.10 & 21.00 & 15.00
& 28.07 & 28.05 & 52.70 \\

Geo2Vec
& \underline{6.40} & \underline{13.00} & \underline{5.40}
& \underline{10.20} & \underline{13.00} & \underline{12.09} \\

\textbf{SpatialVec}
& \textbf{5.20} & \textbf{11.00} & \textbf{4.50}
& \textbf{6.01} & \textbf{5.24} & \textbf{5.75} \\

\bottomrule
\end{tabular}
\vspace{-8pt}
\end{table}

Table~\ref{tab:distance_estimation} shows that SpatialVec achieves the lowest
MAE in all six settings. Relative to Geo2Vec, the strongest baseline, it
reduces the estimation error by $18.8\%$, $15.4\%$, and $16.7\%$ on the
Singapore Pg--Pt, Pl--Pg, and Pg--Pg settings, respectively. The corresponding
reductions on NYC are $41.1\%$, $59.7\%$, and $52.4\%$. The largest gain is
observed for NYC Pl--Pg, where SpatialVec lowers the MAE by approximately
$59.7\%$. \textcolor{black}{This improvement proves SpatialVec's ability to preserve the
relative distances between spatial objects.} THEORY and Hist2Vec mainly capture spatial distributions, while
Poly2Vec emphasizes global frequency-domain structure, making precise
object-to-object separation harder to recover. Geo2Vec performs considerably
better because its SDF field directly represents geometric proximity.
SpatialVec further combines distance, occupancy, and boundary fields with
boundary-focused sampling, allowing the embeddings to retain both the objects'
extent and the local regions that determine their minimum separation.

\subsubsection{Range-Query Selectivity Prediction}

A range query retrieves all spatial objects from a dataset whose geometries
intersect a specified query window. Following the range-query evaluation
setting, we predict the selectivity of a given query window, defined as the
fraction of objects in the dataset that intersect the window. The downstream
model receives the spatial representation of the input data together with the
query-window information and predicts the corresponding range-query
selectivity.

To provide a controlled comparison, all methods use the same datasets, query
windows, training and test splits, and downstream prediction setting. We
evaluate the predicted selectivity using weighted mean absolute percentage
error (WMAPE), where a lower value indicates better performance. The results
are presented in Table~\ref{tab:range_query}. Table~\ref{tab:range_query} shows that SpatialVec achieves the lowest WMAPE
on both datasets. Compared with the strongest competing method, SpatialVec
reduces the prediction error by approximately $11.5\%$ on NYC and $5.7\%$ on Singapore.

\begin{table}[t]
\centering
\caption{Comparing WMAE on range-query's selectivity prediction.}
\label{tab:range_query}

\small
\setlength{\tabcolsep}{6pt}
\renewcommand{\arraystretch}{1.08}

\begin{tabular}{@{}lcc@{}}
\toprule
Method
& NYC$\downarrow$
& Singapore$\downarrow$ \\
\midrule

Hist2Vec
& 0.4448
& \underline{0.2931} \\

Poly2Vec
& \underline{0.4430}
& 0.3130 \\

Geo2Vec
& 0.4562
& 0.3523 \\

\textbf{SpatialVec}
& \textbf{0.3919}
& \textbf{0.2765} \\

\bottomrule
\end{tabular}
\vspace{-8pt}
\end{table}

\subsubsection{Computational Efficiency}

\begin{table}[t]
\centering
\caption{Comparing embedding preprocessing and training latency.}
\label{tab:runtime_training}

\small
\setlength{\tabcolsep}{3.5pt}
\renewcommand{\arraystretch}{1.1}

\begin{tabular}{@{}lccc@{}}
\toprule
Method
& \shortstack{Preprocessing\\Time}
& \shortstack{Training Time}
& \shortstack{Total Time} \\
\midrule

Hist2Vec
& \underline{0.3 min}
& \textbf{250.00 min}
& \textbf{250.3 min} \\

Poly2Vec
& \textbf{0.1 min}
& \underline{300 min}
& \underline{300.01 min} \\

Geo2Vec
& 6.4 min
& 473.4 min
& 479.8 min \\

SpatialVec
& 2.4 min
& 507.6 min
& 510.00 min \\

\bottomrule
\end{tabular}
\vspace{-8pt}
\end{table}

To address RQ2, we evaluate the runtime required to generate spatial
embeddings and examine how it changes with geometric complexity. Table~\ref{tab:runtime_training} compares the embedding generation time of SpatialVec with that of recent baselines which are competitive in prediction and classification tasks.
Although SpatialVec requires more runtime than the baselines, these baselines showed significantly poorer performance in downstream prediction and classification tasks. Since the embedding generation latency is a one-time cost, the superior performance in downstream tasks outweighs the onetime runtime cost.

\begin{figure}[t]
\centering

\begin{tikzpicture}
\begin{axis}[
    width=0.95\columnwidth,
    height=0.65\columnwidth,
    ymode=log,
    xlabel={Number of Vertices},
    ylabel={Latency (ms)},
    xtick={1,2,3},
    xticklabels={$\leq 8$, $9$--$32$, $>32$},
    ymin=0.05,
    ymax=1000,
    grid=major,
    tick label style={font=\scriptsize},
    legend style={
    font=\fontsize{5.5}{6.5}\selectfont,
    at={(0.5,0.98)},
    anchor=north,
    legend columns=4,
    fill=white,
    draw=black,
    column sep=2pt,
    inner xsep=2pt,
    inner ysep=1.5pt
    },
    legend image post style={xscale=0.75}
]

\addplot[
    blue,
    thick,
    mark=*
] coordinates {
    (1,0.078)
    (2,0.082)
    (3,0.100)
};
\addlegendentry{Hist2Vec}

\addplot[
    orange,
    thick,
    mark=*
] coordinates {
    (1,2.35)
    (2,7.96)
    (3,45.27)
};
\addlegendentry{Poly2Vec}

\addplot[
    green!60!black,
    thick,
    mark=*
] coordinates {
    (1,72.4)
    (2,87.5)
    (3,169.0)
};
\addlegendentry{Geo2Vec}

\addplot[
    purple,
    thick,
    mark=*
] coordinates {
    (1,6.85)
    (2,7.43)
    (3,10.54)
};
\addlegendentry{SpatialVec}

\end{axis}
\end{tikzpicture}

\caption{Embedding-generation latency with increasing polygon complexity
on the NYC dataset. The y-axis uses a logarithmic scale.}
\label{fig:runtime_complexity}
\vspace{-6pt}

\end{figure}
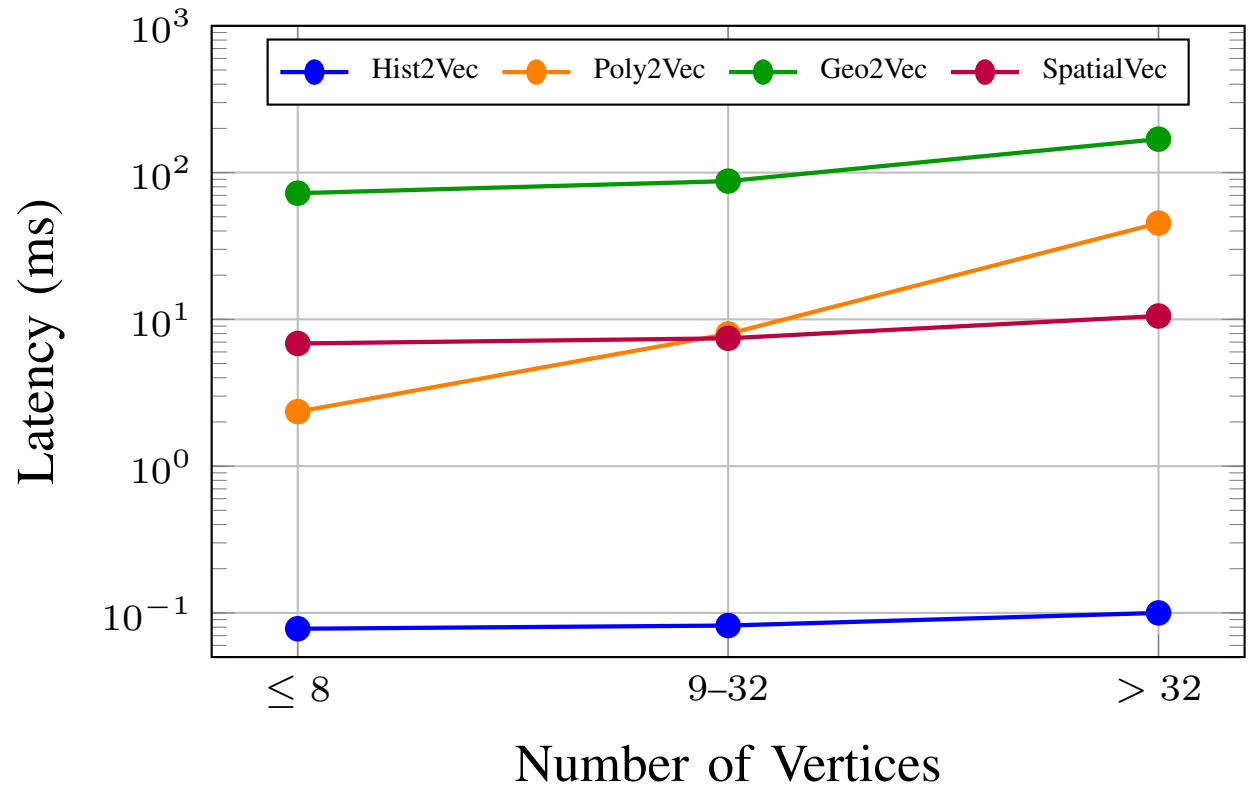

Fig.~\ref{fig:runtime_complexity} further evaluates embedding-generation efficiency by examining how runtime changes as polygon complexity increases.
While Hist2Vec is the fastest for each polygon complexity, SpatialVec maintains relatively
stable embedding-generation latency as polygon complexity increases. Its
runtime grows only modestly across the three complexity groups, whereas
Poly2Vec and Geo2Vec become substantially more expensive for complex polygons.
For polygons in the highest-complexity group, SpatialVec reduces latency by up to $76.7\%$ and $93.8\%$ relative to Poly2Vec and Geo2Vec, respectively.

\subsubsection{Ablation Study}

\begin{table}[t]
\centering
\caption{Accuracy on topological relationship classification on NYC dataset for \textsc{SpatialVec} variants.}
\label{tab:ablation-acc}

\scriptsize
\setlength{\tabcolsep}{6pt}
\renewcommand{\arraystretch}{1.05}

\begin{tabular}{@{}lccccc@{}}
\toprule
Variant
& Pt--Pl
& Pt--Pg
& Pl--Pl
& Pl--Pg
& Pg--Pg \\
\midrule

Full \textsc{SpatialVec}
& \textbf{99}
& \textbf{99}
& \textbf{98}
& \textbf{80}
& \textbf{86} \\

\midrule

\textcolor{black}{w/o boundary-aware sampling}
& 97 & 98 & 95 & 79 & 75 \\

\textcolor{black}{w/o occupancy and boundary fields}
& 97 & 98 & \textbf{98} & 72 & 75 \\

w/o metadata and location integration
& 98 & 98 & 97 & 78 & 76 \\

w/o boundary-aware pooling
& 95 & 97 & 96 & 74 & 75 \\

\textcolor{black}{w/o multi-view aggregation}
& 96 & 98 & 94 & 73 & 76 \\

\bottomrule
\end{tabular}
\vspace{-8pt}
\end{table}

To address RQ3, we conduct an ablation study that evaluates the effectiveness of various key components in our proposed embedding generation technique.
We create multiple variants of SpatialVec by modifying or removing one component of SpatialVec at a time while keeping the remaining components unchanged. For the ``w/o boundary-aware sampling'' variant, we replace the proposed sampling strategy with simple uniform sampling over the canonical space. For the ``w/o occupancy and boundary fields'' variant, the occupancy and boundary values are removed from the implicit field representation while the SDF information is retained. For the ``w/o metadata and location integration'' variant, the metadata and original-location information are removed from the encoder input. For the ``w/o boundary-aware pooling'' variant, the boundary-specific pooling operation is removed while the remaining aggregation views are retained. Finally, for the ``w/o multi-view aggregation'' variant, the proposed multi-view aggregation is replaced with simple mean pooling over all encoded point representations, giving equal importance to samples from all spatial regions.

Table~\ref{tab:ablation-acc} compares all these variants with full SpatialVec on topological classification task using NYC dataset, which shows that the full SpatialVec model achieves the
best overall performance. The largest reduction in accuracy, approximately
$5.4\%$, occurs when boundary-aware pooling or multi-view aggregation is
removed, followed by removing the occupancy and boundary fields ($4.8\%$),
boundary-aware sampling ($3.9\%$), and metadata and location integration
($3.2\%$). The effects are most pronounced for Pg--Pg and Pg--Pl, where
removing the occupancy and boundary fields reduces the mean accuracy by
$11.4\%$, while removing boundary-aware pooling or multi-view aggregation
reduces it by approximately $10.2\%$. These results highlight the importance
of all the components of SpatialVec.
\section{Related Work}
\noindent\textbf{\textit{Graph-Based Spatial Representations.}}
A major line of research in geospatial representation learning focuses on learning compact embeddings for locations, regions, and spatial graphs. PlaceField2BVec \cite{PlaceField2BVec}, Loc2Vec \cite{Loc2Vec}, and Region2Vec \cite{Region2Vec} learn representations from geographic context, spatial proximity, and interactions among predefined locations or regions. Graph-based representation learning methods node2vec \cite{node2vec} and LINE \cite{LINE}
similarly encode entities through graph connectivity and neighborhood structures. These approaches are particularly useful when geographic data can be naturally represented as nodes, edges, or predefined regions. However, their representations cannot explicitly model continuous geometry of spatial objects, including their boundaries, shapes, interiors, and spatial extent, making them less suitable for geometry-sensitive spatial reasoning tasks.\\
\noindent\textbf{\textit{Trajectory-Based Spatial Representations.}}
Another line of work learns spatial representations from movement and trajectory data. ST2Vec \cite{ST2Vec}, contrastive trajectory similarity learning with dual-feature attention \cite{dualattention}, and DeepMove \cite{feng2018deepmove} capture spatial and temporal dependencies from mobility traces. These methods have demonstrated strong performance in applications such as trajectory similarity estimation, mobility prediction, and next-location recommendation. While effective for sequential movement data, they are not designed to represent static geographic objects such as points, polylines, and polygons, which makes them orthogonal to our objectives.
\noindent\textbf{\textit{Geometry-Aware Spatial Representations.}}
More recent spatial representation methods have increasingly focused on
encoding the intrinsic geometry of spatial objects directly from their
coordinate-based representations. Poly2Vec~\cite{poly2vec} encodes points, polylines, and polygons using
multi-frequency Fourier features, while Geo2Vec~\cite{geo2vec} represents
these geometries through sampled SDF field. Both retain overall object
shape, but fine-scale boundary details and localized geometric variations 
receive limited emphasis in the learned representation.

\section{Conclusion}

In this paper, we propose SpatialVec, a unified spatial representation
learning framework for point, polyline, and polygon geometries. SpatialVec
combines canonical normalization, adaptive boundary-aware sampling, implicit
spatial fields, and Transformer-based encoding and decoding to learn fixed-dimensional
embeddings. Its boundary-aware sampling places more observations near
geometric boundaries, while the distance, occupancy, and boundary fields
jointly describe the shape and spatial extent of each object.
Experiments on NYC and Singapore datasets demonstrate that SpatialVec's embedding achieves strong performance on downstream geospatial tasks such as topological relationship classification, distance prediction, and range-query's selectivity prediction and outperforms state-of-the-art baselines by upto 59.7\%. The code and data are available at https://github.com/daiss-lab/SpatialVec.







\bibliographystyle{IEEEtran}
\bibliography{reference}

@inproceedings{appli_urban,
  title={Urban region representation learning with openstreetmap building footprints},
  author={Li, Yi and Huang, Weiming and Cong, Gao and Wang, Hao and Wang, Zheng},
  booktitle={Proceedings of the 29th ACM SIGKDD Conference on Knowledge Discovery and Data Mining},
  pages={1363--1373},
  year={2023}
}

@inproceedings{appli_trans,
  title={M3: Recommendation via Attention-Graph Cluster Q-Learning with Multi-Scale Spatial Heterogeneity for Multi-Purpose, Multi-Stakeholder Green Attractions in Transportation},
  author={Lai, Shih-Yu and Hsieh, Tzu-Hsin and Tsai, Pei-Chi and Kung, Chao-Chun and Ling, Sing-Kai and Hsieh, Hsun-Ping},
  booktitle={Proceedings of the 33rd ACM International Conference on Advances in Geographic Information Systems},
  pages={39--51},
  year={2025}
}

@article{appli_spaitalreason,
  title={SpaRAGraph: Spatial reasoning using retrieval-augmented generation},
  author={Georgiadis, Thanasis and Pavlopoulos, John and Mamoulis, Nikos},
  journal={ACM Transactions on Spatial Algorithms and Systems},
  volume={12},
  number={3},
  pages={1--31},
  year={2026},
  publisher={ACM New York, NY}
}

@article{appli_spaitalreason2,
  title={GS-QA: A Benchmark for Geospatial Question Answering},
  author={Saeedan, Majid and Rashid, Muhammad Shihab and Eldawy, Ahmed and Hristidis, Vagelis},
  journal={arXiv preprint arXiv:2605.22811},
  year={2026}
}

@article{appli_weather,
  title={Context-Aware Multimodal Representation Learning for Spatio-Temporally Explicit Environmental Modelling},
  author={Peters, Julia and Mora, Karin and Mahecha, Miguel D and Ji, Chaonan and Montero, David and Mosig, Clemens and Kraemer, Guido},
  journal={arXiv preprint arXiv:2511.11706},
  year={2025}
}

@article{poly2vec,
  title={Poly2Vec: Polymorphic Fourier-Based Encoding of Geospatial Objects for GeoAI Applications},
  author={Despoina Siampou, Maria and Li, Jialiang and Krumm, John and Shahabi, Cyrus and Lu, Hua},
  journal={arXiv e-prints},
  pages={arXiv--2408},
  year={2024}
}

@inproceedings{geo2vec,
  title={Geo2vec: Shape-and distance-aware neural representation of geospatial entities},
  author={Chu, Chen and Shahabi, Cyrus},
  booktitle={Proceedings of the AAAI Conference on Artificial Intelligence},
  volume={40},
  number={23},
  pages={18985--18993},
  year={2026}
}

@article{PlaceField2BVec,
  title={PlaceField2BVec: A bionic geospatial location encoding method for hierarchical temporal memory model},
  author={Chen, Zugang and Wang, Shaohua and Wu, Kai and Li, Guoqing and Li, Jing and Wang, Jian},
  journal={International Journal of Applied Earth Observation and Geoinformation},
  volume={136},
  pages={104402},
  year={2025},
  publisher={Elsevier}
}

@inproceedings{Loc2Vec,
  title={Location embedding and deep convolutional neural networks for next location prediction},
  author={Sassi, Abdessamed and Brahimi, Mohammed and Bechkit, Walid and Bachir, Abdelmalik},
  booktitle={2019 IEEE 44th LCN symposium on emerging topics in networking (LCN symposium)},
  pages={149--157},
  year={2019},
  organization={IEEE}
}

@inproceedings{Region2Vec,
  title={Region2Vec: community detection on spatial networks using graph embedding with node attributes and spatial interactions},
  author={Liang, Yunlei and Zhu, Jiawei and Ye, Wen and Gao, Song},
  booktitle={Proceedings of the 30th international conference on advances in geographic information systems},
  pages={1--4},
  year={2022}
}

@inproceedings{node2vec,
  title={node2vec: Scalable feature learning for networks},
  author={Grover, Aditya and Leskovec, Jure},
  booktitle={Proceedings of the 22nd ACM SIGKDD international conference on Knowledge discovery and data mining},
  year={2016}
}

@inproceedings{LINE,
  title={Line: Large-scale information network embedding},
  author={Tang, Jian and Qu, Meng and Wang, Mingzhe and Zhang, Ming and Yan, Jun and Mei, Qiaozhu},
  booktitle={Proceedings of the 24th international conference on world wide web},
  pages={1067--1077},
  year={2015}
}

@inproceedings{ST2Vec,
  title={Spatio-temporal trajectory similarity learning in road networks},
  author={Fang, Ziquan and Du, Yuntao and Zhu, Xinjun and Hu, Danlei and Chen, Lu and Gao, Yunjun and Jensen, Christian S},
  booktitle={Proceedings of the 28th ACM SIGKDD conference on knowledge discovery and data mining},
  pages={347--356},
  year={2022}
}

@inproceedings{dualattention,
  title={Contrastive trajectory similarity learning with dual-feature attention},
  author={Chang, Yanchuan and Qi, Jianzhong and Liang, Yuxuan and Tanin, Egemen},
  booktitle={2023 IEEE 39th International conference on data engineering (ICDE)},
  pages={2933--2945},
  year={2023},
  organization={IEEE}
}

@inproceedings{feng2018deepmove,
  title={Deepmove: Predicting human mobility with attentional recurrent networks},
  author={Feng, Jie and Li, Yong and Zhang, Chao and Sun, Funing and Meng, Fanchao and Guo, Ang and Jin, Depeng},
  booktitle={Proceedings of the 2018 world wide web conference},
  year={2018}
}

@article{burns2026community,
  title={Community-engaged dissemination and implementation: restaurant accessibility for people with disabilities},
  author={Burns, Suzanne Perea and Urban, Michael and Sedillo, Jessica Salazar and Frediani, Robert and Samaun, Mohd Anayetul Karim and Tomsicek, Jack and Mendonca, Rochelle},
  journal={OTJR: Occupational Therapy Journal of Research},
  volume={46},
  number={2},
  pages={231--239},
  year={2026},
  publisher={SAGE Publications Sage CA: Los Angeles, CA}
}

@article{theory,
  title={Multi-scale representation learning for spatial feature distributions using grid cells},
  author={Mai, Gengchen and Janowicz, Krzysztof and Yan, Bo and Zhu, Rui and Cai, Ling and Lao, Ni},
  journal={arXiv preprint arXiv:2003.00824},
  year={2020}
}

@article{hist2vec,
  title={A generic machine learning model for spatial query optimization based on spatial embeddings},
  author={Belussi, Alberto and Migliorini, Sara and Eldawy, Ahmed},
  journal={ACM Transactions on Spatial Algorithms and Systems},
  volume={10},
  pages={1--33},
  year={2024},
  publisher={ACM New York, NY}
}

@inproceedings{direct,
  title={Geo-aware networks for fine-grained recognition},
  author={Chu, Grace and Potetz, Brian and Wang, Weijun and Howard, Andrew and Song, Yang and Brucher, Fernando and Leung, Thomas and Adam, Hartwig},
  booktitle={Proceedings of the IEEE/CVF International Conference on Computer Vision Workshops},
  pages={0--0},
  year={2019}
}

@inproceedings{tile,
  title={Birdsnap: Large-scale fine-grained visual categorization of birds},
  author={Berg, Thomas and Liu, Jiongxin and Lee, Seung Woo and Alexander, Michelle L and Jacobs, David W and Belhumeur, Peter N},
  booktitle={2014 IEEE conference on computer vision and pattern recognition},
  pages={2019--2026},
  year={2014},
  organization={IEEE}
}

@inproceedings{warp,
  title={Presence-only geographical priors for fine-grained image classification},
  author={Mac Aodha, Oisin and Cole, Elijah and Perona, Pietro},
  booktitle={Proceedings of the IEEE/cvf international conference on computer vision},
  year={2019}
}

@inproceedings{t2vec,
  title={Deep representation learning for trajectory similarity computation},
  author={Li, Xiucheng and Zhao, Kaiqi and Cong, Gao and Jensen, Christian S and Wei, Wei},
  booktitle={2018 IEEE 34th international conference on data engineering (ICDE)},
  year={2018}
}

@inproceedings{tjepa,
  title={T-jepa: A joint-embedding predictive architecture for trajectory similarity computation},
  author={Li, Lihuan and Xue, Hao and Song, Yang and Salim, Flora},
  booktitle={Proceedings of the 32nd ACM international conference on advances in geographic information systems},
  pages={569--572},
  year={2024}
}

@article{nuftspec,
  title={Towards general-purpose representation learning of polygonal geometries},
  author={Mai, Gengchen and Jiang, Chiyu and Sun, Weiwei and Zhu, Rui and Xuan, Yao and Cai, Ling and Janowicz, Krzysztof and Ermon, Stefano and Lao, Ni},
  journal={GeoInformatica},
  volume={27},
  pages={289--340},
  year={2023},
  publisher={Springer}
}

\end{document}